\documentclass{article}

\usepackage[preprint]{neurips_2025}

\usepackage[utf8]{inputenc} % allow utf-8 input
\usepackage[T1]{fontenc}    % use 8-bit T1 fonts
\usepackage{hyperref}       % hyperlinks
\usepackage{url}            % simple URL typesetting
\usepackage{booktabs}       % professional-quality tables
\usepackage{amsfonts}       % blackboard math symbols
\usepackage{nicefrac}       % compact symbols for 1/2, etc.
\usepackage{microtype}      % microtypography
\usepackage{graphicx}
\usepackage{amsmath}
\usepackage{tabularx}
\usepackage{hhline}
\usepackage{adjustbox}
\usepackage{multirow}
\usepackage{amssymb}     % for \checkmark
\usepackage{pifont}      % for \ding{55}
\usepackage[table]{xcolor}
\usepackage{makecell}

\title{Neural Tangent Knowledge Distillation for Optical Convolutional Networks}

\author{%
  Jinlin Xiang\footnotemark[1]\hspace{0.3em}\footnotemark[3]
  \And
  Minho Choi\footnotemark[1]\hspace{0.3em}\footnotemark[3]
  \And
  Yubo Zhang\footnotemark[3]
  \And
  Zhihao Zhou\footnotemark[3]
  \And
  Arka Majumdar\footnotemark[3]\hspace{0.3em}\footnotemark[4]
  \And
  Eli Shlizerman\footnotemark[2]\hspace{0.3em}\footnotemark[3]\hspace{0.3em}\footnotemark[5]
}

\begin{document}

\maketitle

\footnotetext[1]{These authors contributed equally to this work.}
\footnotetext[2]{Department of Applied Mathematics, University of Washington, Seattle, WA 98103.}
\footnotetext[3]{Department of Electrical and Computer Engineering, University of Washington, Seattle, WA 98103.}
\footnotetext[4]{Department of Physics, University of Washington, Seattle, WA 98103.}
\footnotetext[5]{Corresponding author: shlizee@uw.edu}

\begin{abstract}
Hybrid Optical Neural Networks (ONNs, typically consisting of an optical frontend and a digital backend) offer an energy-efficient alternative to fully digital deep networks for real-time, power-constrained systems. However, their adoption is limited by \textbf{two main challenges}: the accuracy gap compared to large-scale networks during training, and discrepancies between simulated and fabricated systems that further degrade accuracy. While previous work has proposed end-to-end optimizations for specific datasets (e.g., MNIST) and optical systems, these approaches typically lack generalization across tasks and hardware designs. To address these limitations, we propose a task-agnostic and hardware-agnostic pipeline that supports image classification and segmentation across diverse optical systems. To assist optical system design before training, we estimate achievable model accuracy based on user-specified constraints such as physical size and the dataset. For training, we introduce Neural Tangent Knowledge Distillation (NTKD), which aligns optical models with electronic teacher networks, thereby narrowing the accuracy gap. After fabrication, NTKD also guides fine-tuning of the digital backend to compensate for implementation errors. Experiments on multiple datasets (e.g., MNIST, CIFAR, Carvana Masking) and hardware configurations show that our pipeline consistently improves ONN performance and enables practical deployment in both pre-fabrication simulations and physical implementations.

\end{abstract}
\section{Introduction}

\begin{figure}[htbp]
    \centering
    \includegraphics[width=\textwidth]{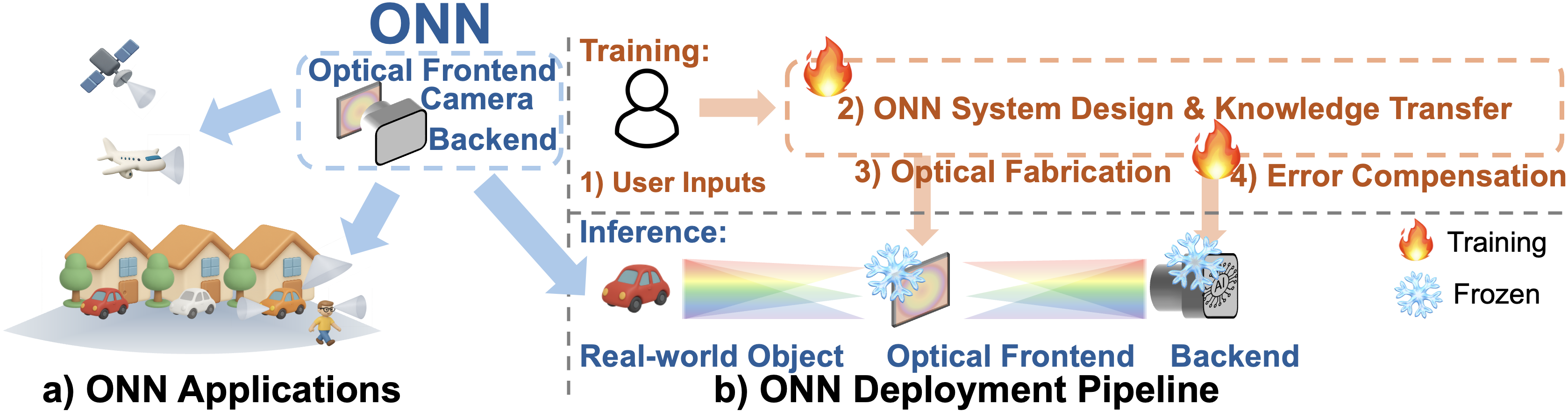}
    \caption{Overview of potential applications for ONNs and our proposed deployment pipeline. (a) ONNs for real-time decision-making in power-constrained scenarios. (b) Our proposed pipeline includes user-driven design, knowledge transfer training, fabrication, and error compensation.}
    \label{fig_teaser}
\end{figure}

Optical Neural Networks (ONNs) offer a promising approach to achieve efficient computation and energy use compared to digital implementations such as Convolutional Neural Networks (CNNs) and Vision Transformers (ViTs), making them well-suited for resource-constrained, real-time physical systems~\cite{xu2021survey,zhao2024towards}. For example, ONNs have been proposed for power-limited applications (illustrated in Figure~\ref{fig_teaser}.a), including satellites~\cite{fu2024optical}, unmanned aerial vehicles~\cite{liu2024towards}, smart home devices~\cite{boloor2025privateeye}, autonomous driving systems~\cite{liu2025extrememeta}, and wearable electronics~\cite{zhao2023integrated}.

Among different ONN implementations, hybrid optical-electronic architectures are practical options under current hardware constraints~\cite{xu2021survey}. In such systems, the optical frontend accelerates computation at the speed of light, while the digital backend refines predictions to improve robustness~\cite{colburn2019optical}. The optical frontend generally consists of a single linear layer (as shown in Figure~\ref{fig_teaser}.a), since: (1) implementing nonlinear activation functions in physical optics remains extremely challenging due to material and device limitations; and (2) without nonlinearity, multiple linear transformations can be mathematically compressed into a single linear transformation. Moreover, from a theoretical standpoint, the universal approximation property (UAP) could still be satisfied by shallow networks, suggesting that hybrid ONNs retain sufficient expressive power for a wide range of tasks when appropriately optimized~\cite{lu2017expressive, zhou2020universality, bao2021approximation}.

Despite their promises, ONNs remain difficult to design and train due to both \textbf{architectural limitations} and \textbf{fabrication-related challenges}. First, existing ONN architectures are typically significantly simpler than modern deep CNNs or ViTs. These simplifications cannot be directly obtained through pruning or quantization~\cite{han2015learning,molchanov2016pruning,seleznova2023neural}.  Second, physical fabrication and experimental deployment inevitably introduce various sources of noise, such as optical misalignment, material variability, and measurement noise, further degrading performance~\cite{liu2024physics}. While some end-to-end optimization strategies have been proposed to address these challenges, they are typically designed for a specific dataset (e.g., MNIST) and tailored to a particular optical system, rather than providing a generalized solution (as also summarized in Related Works). In contrast, we aim to develop a task-agnostic and hardware-agnostic pipeline that can generalize across different datasets and optical hardware setups.

To address these challenges, knowledge transfer, particularly Knowledge Distillation (KD), offers a promising solution by transferring knowledge from pre-trained digital networks to optical models~\cite{hu2021distilling, xiang2022knowledge}. Moreover, recent work shows that successful KD implicitly leads to student-teacher Neural Tangent Kernel (NTK) similarity, where NTK captures how the network’s predictions change with respect to small changes in its parameters~\cite{harutyunyan2023supervision}. As we show here, utilizing NTK for matching is particularly effective for ONNs, as the NTK provides a linear approximation of network behavior, naturally aligning with the linear operations performed by optical systems.

Thus, we propose a Neural Tangent Knowledge Distillation (NTKD) pipeline that generalizes across different optical network designs and datasets to support multiple tasks such as classification and segmentation (also shown in Figure~\ref{fig_teaser}.b). The pipeline starts with specifying the task, the dataset, and the optical structure. Then, NTKD optimization transfers knowledge from digital teacher models to hybrid ONNs by matching their NTKs, effectively transferring the relational structure between classes rather than just matching final predictions. Furthermore, the pipeline compensates for errors introduced during fabrication and experimental deployment by aligning the student's and teacher's NTKs through a small fraction (e.g., 10\%) of real experimental data.

In summary, our contributions are as follows: \\
 $\bullet$ We introduce a Neural Tangent Knowledge Distillation (NTKD) pipeline that supports diverse tasks and optical structures, addressing the challenges of shallow architectures and physical imperfections.\\
$\bullet$  We experimentally validate our pipeline with different ONN implementations on both classification and segmentation tasks, demonstrating its effectiveness through both simulations and fabrications.\\
$\bullet$ We leverage NTK analysis to estimate the achievable accuracy of given hybrid ONNs, providing theoretical guidance on their design and optimization.

\section{Related Works}
\textbf{ONN Tasks and Implementations:} Table~\ref{tab:onn_works} categorizes ONN applications into two tasks (classification and segmentation) and two optical implementations (monochromatic and polychromatic systems). Most previous work focused on monochromatic ONNs for MNIST image classification, including fully optical systems that performed linear transformations~\cite{lin2018diffractive, mengu20diffractive}, physically nonlinear ONNs that used atomic vapors or intensifiers~\cite{yang2023optical, ryou2021free, wang2023image}, and hybrid architectures that combined an optical frontend with a digital backend~\cite{zheng2022meta, zhou2021large, huang2024photonic}. Previous polychromatic ONNs for classification were limited to small datasets such as CIFAR-10, as ONN architectures faced challenges in scaling to complex benchmarks~\cite{xue2024fully, choi2024transferable}. Segmentation tasks are still in the early stages, with a previous study based only on simulation~\cite{liu2025extrememeta}. Our work considers both classification and segmentation tasks. In our pipeline, image reconstruction is implicitly incorporated by encouraging the optical frontend output to align with the simulated result, as the physical output deviates from simulation and requires correction.

\begin{table}[!b]
\centering
\normalsize
\caption{Summary of previous ONN works categorized by task type (classification, segmentation) and implementation level—either simulation only (denoted as \textbf{Sim}) or with physical fabrication and experimental validation (denoted as \textbf{Fab}).}
\label{tab:onn_works}
\resizebox{\linewidth}{!}{%
\begin{tabular}{ll|c|c|c|c}
\toprule
\textbf{ONN Capability} & \textbf{Works} & \makecell{\textbf{Classification} \\ (Sim)} & \makecell{\textbf{Classification} \\ (Fab)} & \makecell{\textbf{Segmentation} \\ (Sim)} & \makecell{\textbf{Segmentation} \\ (Fab)} \\
\midrule
% \multirow{2}{*}{\textbf{Monochromatic}} 
\textbf{Monochromatic}
  & 2018--2025:~\cite{chang2018hybrid,colburn2019optical,yang2023optical,burgos2021design,hu2022high,ryou2021free,wang2023image,zheng2022meta,zhou2021large,lin2018diffractive,mengu20diffractive, liu2024physics, He24pluggable,wei2024spatially,huang2024photonic,zheng2024multichannel,he2024pluggable,wirth2025compressed,wirth2024compressed} 
  & \checkmark & \checkmark & \ding{55} & \ding{55} \\
  % & 2024--Now:~\cite{liu2024physics, He24pluggable,wei2024spatially,huang2024photonic,zheng2024multichannel,he2024pluggable,wirth2025compressed,wirth2024compressed} 
  % & \checkmark & \checkmark & \ding{55} & \ding{55} \\
\midrule
\multirow{3}{*}{\textbf{Polychromatic}} 
  & 2023--2025:~\cite{xue2024fully,huo2023optical,rahman2023time,choi2024transferable, liu2024physics} 
  & \checkmark & \checkmark & \ding{55} & \ding{55} \\
  & ExtremeMETA (2025)~\cite{liu2025extrememeta} 
  & \ding{55} & \ding{55} & \checkmark & \ding{55} \\
  & \cellcolor{gray!15}\textbf{Ours (NTKD)} 
  & \cellcolor{gray!15}\checkmark & \cellcolor{gray!15}\checkmark & \cellcolor{gray!15}\checkmark & \cellcolor{gray!15}\checkmark \\
\bottomrule

\end{tabular}
}
\end{table}

\textbf{Transfer learning for ONNs:}
Transfer learning, particularly Knowledge Distillation (KD), offers a promising solution for transferring knowledge from pre-trained digital networks to optical models~\cite{xiang2022knowledge}. KD minimizes the divergence between a compact student model’s predictions and those of a pretrained teacher model, thereby encouraging the student to actively mimic the teacher’s behavior~\cite{hu2021distilling}.
Beyond conventional KD, recent studies have shown NTK-based approaches to understand knowledge transfer~\cite{lee2020finite}. For example, theoretical insights into KD transfer risk and data efficiency in wide networks have been established through NTK analysis~\cite{ji2020knowledge}. Subsequent work demonstrated that successful knowledge distillation implicitly led to student-teacher NTK alignment~\cite{harutyunyan2023supervision}, and NTK similarity was further applied to quantify task affinities in multi-task learning~\cite{morello2024exploring}. In contrast, our work targets physically constrained ONNs, and introduces an explicit NTK matching strategy to directly guide the distillation process from a digital teacher to an optical student.

\textbf{Compensation Strategies for Practical ONNs:}
Fabrication imperfections and system noise in physical ONNs often lead to significant performance drops compared to simulations. Some used deep learning to model the system directly in a data-driven manner~\cite{zheng2023dual}, including ONN auto-learning~\cite{spall2022hybrid, ruiz2020deep}, where ONNs were trained to fit experimental input-output mappings. Other methods introduced physical information via hardware-in-the-loop training. For example, physics-constrained frameworks embedded fabrication-aware models and losses to better align learning with optical behavior~\cite{liu2024physics}. Moreover, some approaches avoided simulation entirely by randomly fabricating optical kernels and training a digital backend to adapt to the fixed frontend structure~\cite{zhang2024general,zhao2024frequency,pors2016random,dupre2018design,wray2020light}. This simplified fabrication but placed the learning burden entirely on the backend. It is also not clear if such random surfaces perform better than an ordinary lens. In contrast, our work identifies sources of physical errors and introduces an NTK alignment strategy for effective compensation.

\begin{figure}[!t]
    \centering
    \includegraphics[width=\textwidth]{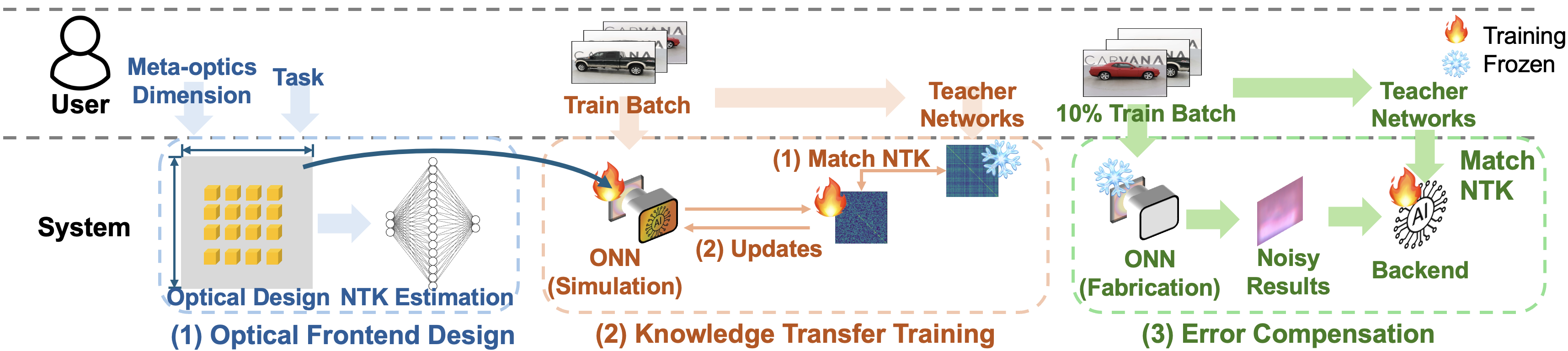}
    \caption{Overview of the pipeline. It consists of three steps: (1) Optical Frontend Design based on user-specified inputs, (2) Knowledge Transfer Training using Neural Tangent Kernel (NTK) matching, and (3) Error Compensation for fabricated optical frontends.}
    \label{fig_pipeline}
\end{figure}

\section{Methods}

\subsection{Optical Frontend Design}
At the initialization of the pipeline, user inputs are required to define the optical system (also shown in Figure~\ref{fig_pipeline}.1). Specifically, the user specifies (1) the physical size of the optical frontend (e.g., the number of meta-optic kernels), (2) the target dataset for the task, and (3) the desired network structure, such as the number of layers and channels. The optical convolution is realized either through a 4f system or via point spread function (PSF)-based free-space propagation~\cite{choi2024transferable,chang2018hybrid}.

\textbf{Optical Frontend Layout:} We consider a metasurface of size ($h,w$), onto which we aim to place $n_\textit{kernels}$ square optical kernels, each of size $k$ (in mm), with a minimum edge-to-edge spacing $d$ to satisfy fabrication constraints. To compute the maximum number of kernels that can be placed while preserving symmetry, we define

\begin{equation}
    n_{\text{cols}} = \left\lfloor \frac{w - d}{k+d} \right\rfloor, \quad
n_{\text{rows}} = \left\lfloor \frac{h - d}{k+d} \right\rfloor, \quad
n_\textit{kernels} = n_{\text{cols}} \times n_{\text{rows}}.
\end{equation}

\textbf{Performance Estimation:} 
Once the physical layout of the ONN is determined, we aim to estimate its expected performance without empirical training. We adopt the Neural Tangent Kernel (NTK) framework, which captures the training dynamics of \textbf{infinitely wide neural networks} under gradient descent. In particular, we introduce a reference network that shares the same architecture as the designed ONN (e.g., number of layers and connectivity) but has infinite width at each layer. Under this assumption, the predictions of the reference network correspond to NTK regression~\cite{jacot2018neural,NEURIPS2019_dbc4d84b}. Let the reference network $f(x; \theta)$ be a neural network parameterized by $ \theta $, which maps an input $x$ to an output $f(x; \theta) $. The NTK is defined as
\begin{equation}
    \Theta(x, x') = \nabla_\theta f(x; \theta)^\top \nabla_\theta f(x'; \theta),
\end{equation}
where $\nabla_\theta f(x; \theta)$ is the Jacobian of the network output with respect to its parameters. Let $\{x_i^{\text{train}}, y_i^{\text{train}}\}_{i=1}^{n_{\text{train}}}$ be the training data and $\{x_i^{\text{test}}\}_{i=1}^{n_{\text{test}}}$ be the test data. We compute
\begin{equation}
\Theta_{\text{train,train}} = \Theta(x^{\text{train}}, x^{\text{train}}) \in \mathbb{R}^{n_{\text{train}} \times n_{\text{train}}}, \quad
\Theta_{\text{test,train}} = \Theta(x^{\text{test}}, x^{\text{train}}) \in \mathbb{R}^{n_{\text{test}} \times n_{\text{train}}}.
\end{equation}
and use kernel regression to predict outputs on the test set
\begin{equation}
   f(x^{\text{test}};\theta) = \Theta_{\text{test,train}} \left( \Theta_{\text{train,train}} + \lambda I \right)^{-1} y^{\text{train}}, 
\end{equation}
where $\lambda$ is a regularization parameter, which is selected via grid search on a validation set.

The NTK-based performance estimation serves as a diagnostic tool to evaluate whether the specified ONN architecture is expressive enough for the given task. While this estimation is not used for training or loss computation, it provides an early signal to guide architectural decisions and allows users to iteratively refine the optical design before full training and fabrication. For example, if the estimated test accuracy is much lower than the expected performance, it may suggest a mismatch between the ONN's capacity (e.g., depth) and task complexity.

\subsection{Knowledge Transfer Training}\label{train}

After the user specifies the ONN architecture, we train the system for the target task. We define a supervised learning problem with input-output pairs \((x, y)\), where \(x\) represents the input samples and \(y\) denotes the corresponding ground-truth labels. The network parameters ($\theta$), including the optical frontend and the digital backend, are initialized and optimized jointly (shown in Figure~\ref{fig_pipeline}.2).

\textbf{End-to-end loss:} The first loss term we consider is a standard end-to-end supervision loss, which directly minimizes the discrepancy between the network’s predictions and the ground-truth labels. Formally, we optimize the following objective
\begin{equation}
\mathcal{L}_{\text{E2E}} = \mathbb{E}_{(x, y) \sim \mathcal{D}} \left[ \ell(f_\textit{ONN}(x; \theta), y) \right],
\end{equation}
where \(\ell(\cdot, \cdot)\) is a standard loss function such as cross-entropy for classification tasks, \(f_\textit{ONN}(x; \theta)\) denotes the output of the network with parameters \(\theta\), and \(\mathcal{D}\) represents the training dataset.

\textbf{Neural Tangent Knowledge Distillation (NTKD) Loss:} In addition to the standard end-to-end loss, we introduce a knowledge transfer loss based on Neural Tangent Kernel (NTK). Specifically, we assume access to a pretrained teacher network, such as LeNet for MNIST, AlexNet for CIFAR-10, or U-Net for image segmentation tasks.

Given a minibatch of input samples \( \{x_i\}_{i=1}^{n_{\text{batch}}} \), we compute the Jacobian matrices of both the teacher network ($f_{\text{teacher}}$) and student ONN ($f_{\text{ONN}}$) with respect to their parameters ($\theta_{\text{teacher}}, \theta_{\text{ONN}}$)
\begin{equation}
J_{\text{teacher}} = \left[ \frac{\partial f_{\text{teacher}}(x_i)}{\partial \theta_{\text{teacher}}} \right]_{i=1}^{n_{\text{batch}}}, \quad
J_{\text{ONN}} = \left[ \frac{\partial f_{\text{ONN}}(x_i)}{\partial \theta_{\text{ONN}}} \right]_{i=1}^{n_{\text{batch}}}.
\end{equation}
The Jacobians \( J_{\text{teacher}} \in \mathbb{R}^{n_{\text{batch}} \times p_{\text{teacher}}} \) and \( J_{\text{ONN}} \in \mathbb{R}^{n_{\text{batch}} \times p_{\text{ONN}}} \) may differ in width depending on the number of trainable parameters in each network. Here, \(p_{\text{teacher}}\) and \(p_{\text{ONN}}\) denote the number of parameters in the teacher and ONN, respectively. Their corresponding NTK matrices,
\begin{equation}
\Theta_{\text{teacher}} = J_{\text{teacher}} J_{\text{teacher}}^\top, \quad
\Theta_{\text{ONN}} = J_{\text{ONN}} J_{\text{ONN}}^\top,
\label{eq_ntk}
\end{equation}
are both of size \( n_{\text{batch}} \times n_{\text{batch}} \). We define the NTKD loss by minimizing the discrepancy between the NTK matrices of the teacher network and the ONN (e.g., MSE)
\begin{equation}
\mathcal{L}_{\text{NTKD}} = \mathbb{E}_{\{x_i\}_{i=1}^{n_{\text{batch}}} \sim \mathcal{D}} \left[ \ell\left( \Theta_{\text{teacher}}, \Theta_{\text{ONN}} \right) \right].
\end{equation}
Then, we minimize a weighted sum of two losses, controlled by hyperparameters \(\alpha\) and \(\beta\)
\begin{equation}
\min_{\theta} \left( \alpha \mathcal{L}_{\text{E2E}} + \beta \mathcal{L}_{\text{NTKD}} \right).
\label{eq_total_loss}
\end{equation}

\subsection{Error Compensation}

The physical fabrication uses the optimized simulation parameters obtained through the process described in Section~\ref{train}. The fabrication fixes the optical frontend, and only the digital backend remains tunable. Due to unavoidable fabrication and experimental errors, discrepancies arise between the designed and realized optical system. Given an input image $a$ and convolution kernel $k$, the ideal output is $y = a * k$. The fabricated optical convolution output with noise at location $(i, j)$ is
\begin{equation}
\tilde{y}_{i,j} = \alpha \beta \sum_{m=1}^{k_\textit{size}} \sum_{n=1}^{k_\textit{size}} a_{i+m-1, j+n-1} \left(k_{m,n} + \delta_{m,n}\right) + \epsilon_{i,j}.
\end{equation}

Here, the scaling factors $\alpha$ (image brightness) and $\beta$ (image–kernel misalignment) can be experimentally calibrated to match the designed system, while the sensor noise $\epsilon$ is primarily determined by the imaging device characteristics. We further quantify the impact of fabrication noise $\delta$ (with proof provided in the Supplementary). In particular, we show that perturbations in the NTK caused by kernel fabrication errors (\(\delta_{ij}\)) scale as
\begin{equation}
    \|\Delta \Theta_{\text{ONN}} \| \sim O\left(\frac{\|\delta\|}{m}\right),
    \label{eq_ntk_bound}
\end{equation}
where $\Delta \Theta_{\text{ONN}}$ denotes the NTK perturbation, and $m$ is the number of kernels (i.e., the network width). This result indicates that networks with more kernels are inherently more robust to fabrication noise. Indeed, if $m \gg \|\delta\|$, the impact of fabrication noise on the prediction can be considered negligible. If $\delta$ is interpreted as a gradient descent step, Eq.~\ref{eq_ntk_bound} is consistent with NTK theory: as $m \to \infty$, the NTK ($\Theta$) remains constant during training, implying $\Delta \Theta \to 0$~\cite{jacot2018neural}.

To correct these errors, we re-apply minimization in Eq.~\ref{eq_total_loss}, but restrict optimization to the unfrozen backend parameters (shown in Figure~\ref{fig_pipeline}.3). The teacher network takes the raw input images and computes its NTK matrix over a batch of samples, as defined in Eq.~\ref{eq_ntk}. The student network receives feature maps from the fixed optical frontend, which processes the same batch of input images and produces an NTK matrix of size \( n_{\text{batch}} \times n_{\text{batch}} \).

\section{Results}
\begin{figure}[!t]
    \centering
    \includegraphics[width=\textwidth]{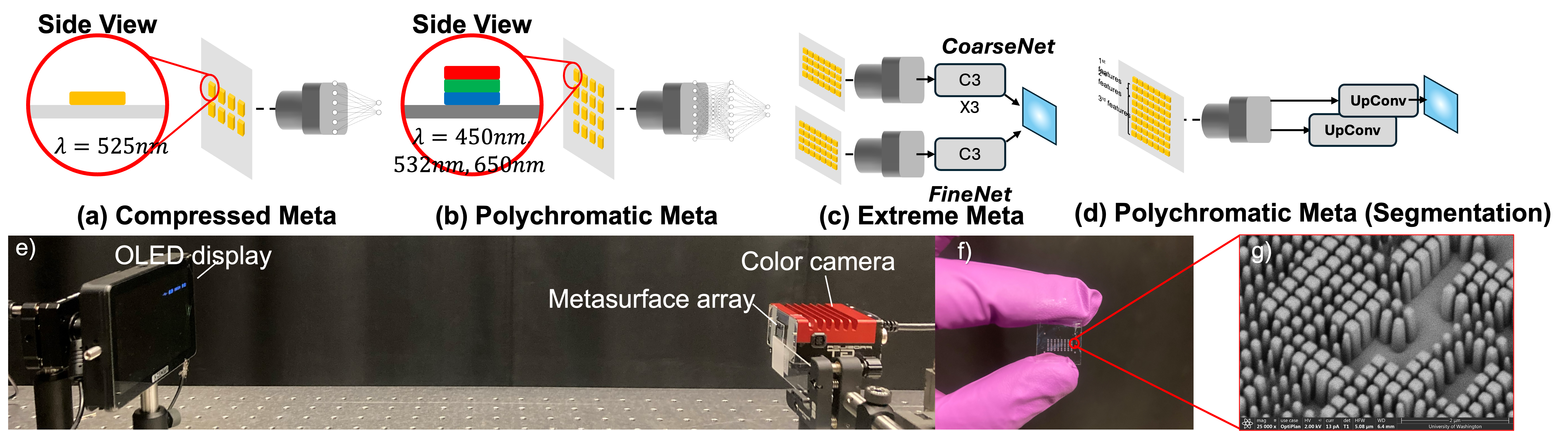}
    \caption{Optical systems. (a) Compressed Meta ONN on MNIST~\cite{wirth2025compressed}; (b) Polychromatic Meta ONN on CIFAR-10~\cite{choi2024transferable}; (c) ExtremeMETA for segmentation with dual optical frontends~\cite{liu2025extrememeta}; (d) Customized Polychromatic Meta ONN for segmentation (Ours); (e) Optical measurement setup; (f) Fabricated PSF-engineered meta-optics; (g) Scanning electron microscopy image of the meta-optics.}
    \label{fig_sys}
\end{figure}

\subsection{Optical Settings, Pre-trained teachers, Datasets and Evaluation Metrics}

\textbf{Optical Settings: } Figure~\ref{fig_sys} demonstrates four different optical systems used in the experiments. For monochromatic image classification, we conducted experiments on the Compressed Meta ONN architecture~\cite{wirth2025compressed} using the MNIST dataset~\cite{lecun1998gradient}. This system consists of a single optical frontend with 8 kernels ($7\times7$) and a compact digital backend composed of two fully connected layers. For polychromatic image classification, we evaluated the Polychromatic Meta ONN~\cite{choi2024transferable} on the CIFAR-10 dataset~\cite{krizhevsky2009learning}. This model consists of a single optical frontend with 16 kernels ($7\times7$) and a digital backend consisting of three fully connected layers.

For image segmentation, we performed experiments on both the Extreme Meta ONN~\cite{liu2025extrememeta} and a modified version of the Polychromatic Meta ONN~\cite{choi2024transferable}, using Kaggle’s Carvana Image Masking dataset. ExtremeMETA consists of two parallel polychromatic optical frontends, followed by a dual-path digital backend composed of CoarseNet for global feature extraction and FineNet for detail enhancement, with their outputs fused to produce the final segmentation. Our customized polychromatic segmentation system extends the Polychromatic Meta ONN by incorporating a single optical frontend with 56 kernels (8, 16, and 32 kernels to capture hierarchical depth representations, each of size $3\times3$) and a backend composed of upconvolutional layers to support dense prediction tasks. For a fair comparison between the two systems, we matched the number of optical kernels. Optical implementation details are in the Supplementary.

In experiments, we manipulate polychromatic$-$red, green, and blue$-$point spread functions (PSFs) of the meta-optics using a gradient descent algorithm. Physical shapes and dimensions of the PSF-engineered meta-optics are shown in Figure~\ref{fig_sys}. Since the meta-optics are designed with PSFs that function as convolutional kernels, optical convolution occurs naturally during image capture. Our experimental setup is straightforward: we replace a conventional imaging lens with PSF-engineered meta-optics. The meta-optics are carefully positioned and aligned with the color camera, enabling the capture of PSFs and convolved images using a laser pointer and an OLED display, respectively. A schematic representation and a photograph of the setup are provided in Figure~\ref{fig_sys}.

\textbf{Pre-trained teachers and Datasets:} The MNIST and CIFAR-10 datasets each consist of $50,000$ training images and $10,000$ testing images. The Carvana dataset, originally introduced in Kaggle’s Carvana Image Masking Challenge, contains 5,088 high-resolution $1920 \times 1280$ car images. We adopt LeNet (99.1\% accuracy on MNIST), AlexNet (84.5\% on CIFAR-10), and a full U-Net (95.4\% mIoU on Segmentation) as teacher models for their respective tasks.

\textbf{Evaluation Metrics:}
For classification tasks, we ran each experiment five times with different random seeds and reported the mean and standard deviation (std) of the classification accuracy. For segmentation tasks, we similarly conducted five independent runs and reported the mean Intersection over Union (mIoU) along with the standard deviation. 

\subsection{Main Results}
\begin{table}[!b]
\centering
\caption{Performance comparison of ONN methods across different tasks and training strategies}
\label{tab:onn_performance}
\resizebox{\linewidth}{!}{%
\begin{tabular}{lcc|cc|cc|cc}
\toprule
\textbf{Methods} 
& \multicolumn{2}{c|}{\textbf{Monochromatic Classification (\%)}} 
& \multicolumn{2}{c|}{\textbf{Polychromatic Classification (\%)}} 
& \multicolumn{4}{c}{\textbf{Polychromatic Segmentation (mIoU)}} \\
% \textbf{ONNs}
& \multicolumn{2}{c|}{Compressed Meta (2025)~\cite{wirth2025compressed}} 
& \multicolumn{2}{c|}{Polychromatic Meta (2025)~\cite{choi2024transferable}} 
& \multicolumn{2}{c}{Extreme Meta (2025)~\cite{liu2025extrememeta}} 
& \multicolumn{2}{c}{Polychromatic Meta (Ours)} \\
\midrule
% & \multicolumn{2}{c}{Accuracy (\%)} & \multicolumn{2}{c|}{Accuracy (\%)} & \multicolumn{2}{c}{mIoU} & \multicolumn{2}{c}{mIoU} \\
% \cmidrule(lr){2-5}
% \cmidrule(lr){6-9}
% \cmidrule(lr){6-7}
% \cmidrule(lr){8-9}

% NTK Estimation ($\infty$-width)         
% & \multicolumn{2}{c}{97.7\%}  & \multicolumn{2}{c|}{67.3\%}   & \multicolumn{2}{c}{0.721} & \multicolumn{2}{c}{0.759} \\
Simulation, No Transfer       
& \multicolumn{2}{c|}{91.4 $\pm$ 0.8\%}  & \multicolumn{2}{c|}{56.4 $\pm$ 1.9\%}  & \multicolumn{2}{c}{68.3 $\pm$ 0.5\%} & \multicolumn{2}{c}{74.3 $\pm$ 0.4\%} \\
Simulation, KD               
& \multicolumn{2}{c|}{95.9 $\pm$ 0.6\%}  & \multicolumn{2}{c|}{72.5 $\pm$ 2.1\%}  & \multicolumn{2}{c}{75.3 $\pm$ 0.2\%} & \multicolumn{2}{c}{86.7 $\pm$ 0.4\%} \\
Simulation, NTKD           
& \multicolumn{2}{c|}{97.3 $\pm$ 0.6\%}  & \multicolumn{2}{c|}{75.6 $\pm$ 0.9\%}  & \multicolumn{2}{c}{80.1 $\pm$ 0.2\%} & \multicolumn{2}{c}{91.5 $\pm$ 0.4\%} \\
\bottomrule
\end{tabular}
}
\end{table}

\begin{figure}[!t]
    \centering
    \includegraphics[width=\textwidth]{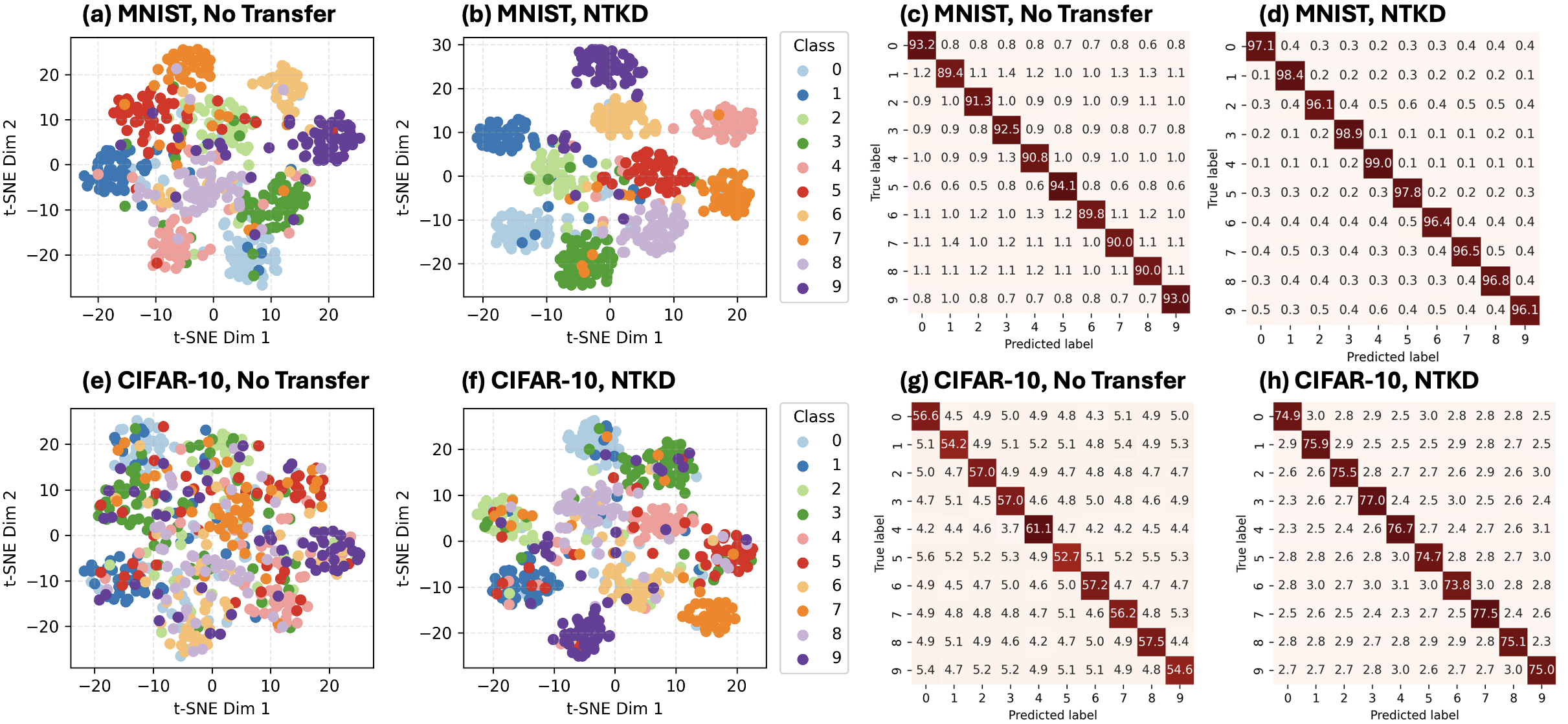}
    \caption{Simulation: T-SNE and confusion matrices in MNIST (a-d) and CIFAR-10 (e-h).}
    \label{Fig_vis_sim}
\end{figure}

\textbf{Simulation Results:} Table \ref{tab:onn_performance} summarizes the accuracy of different ONN training strategies in classification and segmentation tasks. For monochromatic classification, NTKD achieved 97.3\% accuracy and outperformed both KD-based transfer (95.9\%) and the non-transfer baseline (91.4\%). Similar trends were observed in the more challenging polychromatic classification setting, where NTKD achieved 75.6\%, surpassing KD (72.5\%) and baseline (56.4\%). For segmentation tasks, we evaluated models on both the Extreme Meta and our proposed Polychromatic Meta datasets. NTKD consistently achieved higher mIoU scores across both optical systems, surpassing KD-based transfer and end-to-end training without transfer (75.3\% and 86.7\%, respectively). 

These results indicate that training ONNs end-to-end without transfer often leads to suboptimal performance. Incorporating knowledge transfer through KD or NTKD improves learning efficacy and overall segmentation quality. In particular, the NTKD approach outperformed KD in different tasks, demonstrating its ability to guide representation learning in optical neural networks.

Figure~\ref{Fig_vis_sim} demonstrates knowledge transfer strategies on MNIST and CIFAR-10 representations and classification performance. Compared to the no-transfer baseline, these strategies improve class separability in t-distributed Stochastic Neighbor Embedding (t-SNE) visualizations and reduce noise in confusion matrices. NTKD transfer shows improved clustering and accuracy in experiments.

\begin{table}[!b]
\centering
\caption{Evaluation of ONN error compensation methods on fabricated optical systems across both classification and segmentation tasks.}
\label{tab:onn_compensation}
\resizebox{\linewidth}{!}{%
\begin{tabular}{lcc|cc|cc}
\toprule
\textbf{Method} 
& \multicolumn{2}{c|}{\textbf{Monochromatic Classification (\%)}} 
& \multicolumn{2}{c|}{\textbf{Polychromatic Classification (\%)}} 
& \multicolumn{2}{c}{\textbf{Polychromatic Segmentation (mIoU)}}\\
& \multicolumn{2}{c|}{Compressed Meta (2025)~\cite{wirth2025compressed}}
& \multicolumn{2}{c|}{Polychromatic Meta (2025)~\cite{choi2024transferable}}
& \multicolumn{2}{c}{Polychromatic Meta (Ours)} \\
\midrule
% & \multicolumn{2}{c|}{Accuracy(\%)} & \multicolumn{2}{c|}{Accuracy(\%)} & \multicolumn{2}{c}{mIoU}\\
No compensation (baseline)         
& \multicolumn{2}{c|}{89.2\%}  & \multicolumn{2}{c|}{47.3\%}   & \multicolumn{2}{c}{49.7\%}      \\
Error compensation (End-to-End) & \multicolumn{2}{c|}{93.2 $\pm$ 0.1\% }  & \multicolumn{2}{c|}{70.4 $\pm$ 2.1\%}   & \multicolumn{2}{c}{62.7 $\pm$ 0.9\%}\\
Error compensation (NTKD) & \multicolumn{2}{c|}{95.1 $\pm$ 0.1\%}  & \multicolumn{2}{c|}{74.9 $\pm$ 1.3\%}   & \multicolumn{2}{c}{81.2 $\pm$ 0.6\%}\\
\bottomrule
\end{tabular}
}
\end{table}

\begin{figure}[!t]
    \centering
    \includegraphics[width=\textwidth]{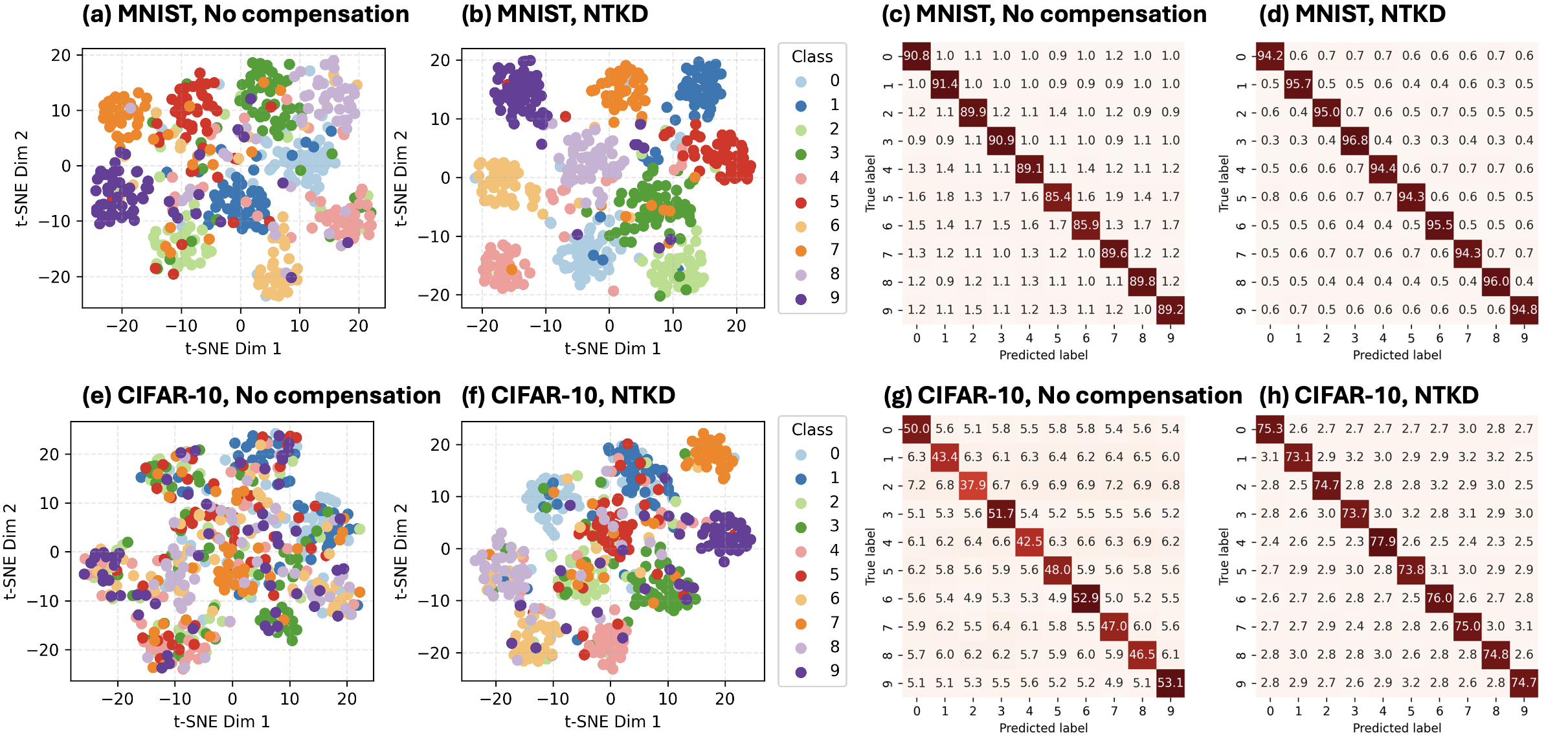}
    \caption{Fabrication: T-SNE and confusion matrices on MNIST (a–d) and CIFAR-10 (e–h).}
    \label{Fig_vis_fab}
\end{figure}

\textbf{Fabrication and Compensation Results:}  Table~\ref{tab:onn_compensation} summarizes the impact of error compensation strategies on ONN performance in classification and segmentation tasks. Due to unavoidable fabrication and experimental errors, optical frontends suffer significant accuracy drops, especially in the polychromatic setting. Without compensation, the monochromatic system exhibits an 8.1\% drop, and the more fabrication-sensitive polychromatic system shows a 28.3\% drop on CIFAR-10 and a 41.8\% reduction in mIoU on the Carvana segmentation task. This performance gap highlights the increased challenge of fabricating RGB-sensitive kernels in polychromatic ONNs compared to monochromatic ONNs. Our results demonstrate that knowledge transfer methods are able to assist with denoising that gap. NTKD compensation yields higher accuracy in both cases (95.1\% for monochromatic, 74.9\% for polychromatic and 81.2\% for image segmentation task), outperforming end-to-end deep learning compensation, and validating its effectiveness in robust corrections for fabrication.

Figure~\ref{Fig_vis_fab} compares the t-SNE visualizations and confusion matrices of MNIST and CIFAR-10 representations under different compensation strategies. Without compensation (Figures~\ref{Fig_vis_fab} a, c, e, g), both optical systems exhibit class overlap in the feature space and reduced classification accuracy, primarily due to fabrication errors and optical misalignments. The NTKD correction (Figures~\ref{Fig_vis_fab} b, d, f, h) compensates for these errors and improves the clustering structure and classification accuracy, demonstrating robustness and generalization across datasets and optical systems.

\subsection{Discussion and Ablation Study}

\textbf{Backend Complexity of ONNs: } The complexity of the backend plays a critical role in the performance of optical systems, particularly in hybrid optical-digital architectures. When a strong digital backend is employed, it can effectively denoise and recover accurate outputs, even when the optical frontend introduces significant noise. However, a strong backend not only diminishes the contribution of the optical frontend but also significantly increases power consumption—undermining the core motivation for adopting optical computing in resource-constrained environments. In such scenarios,  digital networks (e.g., ViT or U-Net) are often a more practical and effective choice than hybrid ONNs with disproportionately strong backends. For practical deployment, we need to carefully balance the computational load between optics and computational backend and find a tradeoff between acceptable energy consumption/ latency and acceptable accuracy, which will be an application dependent trade-off.

\begin{table}[!t]
\centering
\caption{Random PSF kernel design across different tasks}
\label{ablation_random_psf}
\resizebox{\linewidth}{!}{%
\begin{tabular}{lcc|cc|cc}
\toprule
% \textbf{Method} 
& \multicolumn{2}{c|}{\textbf{Monochromatic Classification (\%)}} 
& \multicolumn{2}{c|}{\textbf{Polychromatic Classification (\%)}} 
& \multicolumn{2}{c}{\textbf{Polychromatic Segmentation (mIoU)}}\\
\midrule
% & \multicolumn{2}{c|}{Accuracy(\%)} & \multicolumn{2}{c|}{Accuracy(\%)} & \multicolumn{2}{c}{mIoU}\\
8 kernels         
& \multicolumn{2}{c|}{96.12\%}  & \multicolumn{2}{c|}{36.73\%}   & \multicolumn{2}{c}{49.3\%}      \\
500 kernels  & \multicolumn{2}{c|}{96.81\% }  & \multicolumn{2}{c|}{49.32\%} & \multicolumn{2}{c}{64.1\%}  \\
1000 kernels  & \multicolumn{2}{c|}{97.24\% }  & \multicolumn{2}{c|}{56.23\%}   & \multicolumn{2}{c}{69.9\%}\\
$\infty$ kernels  & \multicolumn{2}{c|}{97.72\%}  & \multicolumn{2}{c|}{67.33\%}   & \multicolumn{2}{c}{72.1\%} \\
% Ours (8 kernels + NTK) & \multicolumn{2}{c|}{96.21\%}  & \multicolumn{2}{c|}{52.32\%}   & \multicolumn{2}{c}{0.592}\\
\bottomrule
\end{tabular}
}
\end{table}

\textbf{Random vs. Designed Parameters: } Another possible direction for designing and training ONNs is to use randomly initialized optical parameters while training only the digital backends. This approach aims to avoid the need for extensive simulation and hardware-in-the-loop optimization of the optical frontends. We conducted experiments using a single optical convolutional layer and a lightweight backend—consisting of a single fully connected layer for classification, or a single upsampling layer for segmentation. As shown in Table~\ref{ablation_random_psf}, increasing the number of random kernels consistently improves performance across tasks. For example, in polychromatic classification, accuracy improves from 36.73\% (8 kernels) to 56.23\% (1000 kernels), and further to 67.33\% in the NTK regime, which approximates an infinite number of random kernels. Similarly, in polychromatic segmentation, mIoU rises from 49.3\% to 69.9\% and reaches 72.1\% under NTK estimation. These results demonstrate that while increasing the number of random PSFs improves performance, they still underperform our approach (designed kernels with knowledge transfer).

\textbf{Scalability of ONNs: } Scaling current ONNs remains challenging, as most designs rely on shallow structures with limited linear computational capacity. Implementing nonlinear operations in ONNs is especially difficult due to physical constraints, such as the limited pixel size. These hardware limitations make it hard for ONNs to support deep and expressive architectures like those used in digital networks. We observed that using different kernels to simulate multiple layers of a digital network leads to better performance, compared to simply compressing a deep CNN into a single-layer ONN. To test the scalability of ONNs under these limitations, we evaluated a polychromatic ONN on the ImageNet-100 dataset and achieved a top-1 accuracy of 46.32\%. In summary, scalable and expressive ONNs will ultimately rely on physical advances enabling deep and nonlinear optical computations. To that end, image intensifiers present an interesting opportunity for cascading multiple metasurfaces, as they can provide both nonlinear activation and signal regeneration~\cite{wang2023image}.

\begin{figure}[!t]
    \centering
    \includegraphics[width=\textwidth]{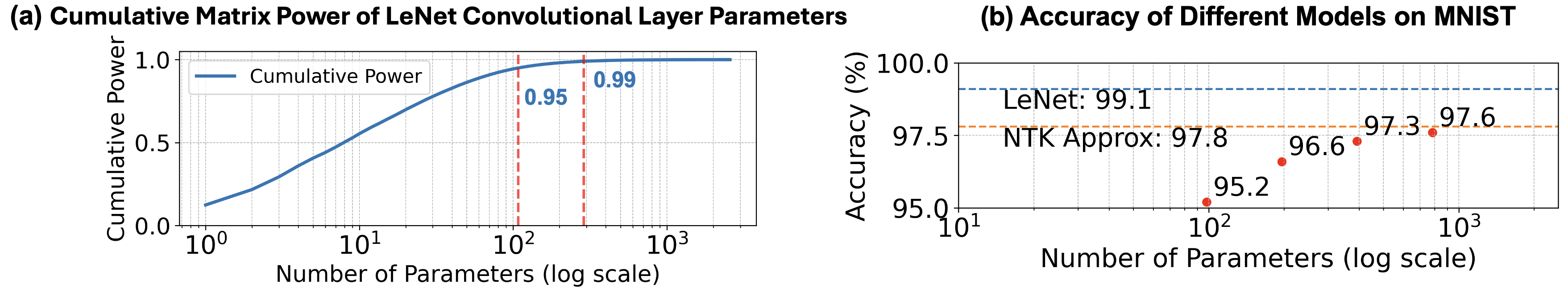}
    \caption{\textbf{NTK compressibility analysis.}
    (a) Cumulative eigenvalue power of the Parameter matrix \(g = J^\top J\) for LeNet's convolutional layers, showing that 95\% and 99\% of power are concentrated in just 108 and 288 parameters.
    (b) Accuracy of models with varying parameter counts.}
    \label{Fig_vis_ntk}
\end{figure}

\textbf{Parameter-Space Compressibility: }   
To examine redundancy of the convolutional layer, we compute the Jacobian \(J \in \mathbb{R}^{n \times p}\) of LeNet outputs with respect to its 2,572 convolutional parameters. We then form the parameter Gram matrix \(J^\top J \in \mathbb{R}^{p \times p}\) and analyze its cumulative eigenvalue spectrum to quantify the parameter effective dimensions.
Figure~\ref{Fig_vis_ntk}.a plots the cumulative eigenvalue power of \(J^\top J\). 108 parameters account for 95\% of the total power, and 288 parameters account for 99\%, indicating strong compressibility.  
To validate this, we trained compressed models with varying parameter counts (Figure~\ref{Fig_vis_ntk}.b), and analyzed the trade-off between model size and accuracy. For example, a model with just 98 parameters in the convolutional layer achieved 95.2\% accuracy, closely matching the 95\% cumulative NTK power threshold. Notably, when the cumulative matrix power exceeded 99\%, further increases in parameter count without changes to the model structure yielded diminishing returns. For example, increasing parameters from 98 to 784—twice that of Compressed Meta~\cite{wirth2024compressed}—only improved accuracy by 0.5\%. This marginal gain did not surpass the accuracy predicted by NTK analysis, highlighting the inefficiency of overparameterization and suggesting the use of dimensionality reduction. The Gram matrix can help estimate the efficient number of parameters for a given task.

\textbf{Fabrication Analysis: }  Several factors may help explain the discrepancy between the designed and measured kernels. First, the local periodic approximation — which simplified the metasurface optics (MO) design by assuming the scatterers were arranged periodically — neglected the coupling between adjacent dissimilar scatterers and could introduce phase errors. Second, unavoidable fabrication errors further contributed to the observed discrepancy. Third, the pre-designed kernel needed to be properly matched to the sensor’s pixel array; any misalignment between the metasurface optics and the camera could also lead to deformation of the measured kernels. Regarding the polychromatic versus single-wavelength kernels, metasurfaces inherently suffered from strong chromatic aberrations, a universal characteristic of diffractive optics. While we co-optimized the kernel across multiple wavelengths during the design process to ensure consistent behavior, the performance remained limited by the intrinsic material properties.

\textbf{MACs and Power Consumption: } We estimated the multiply–accumulate operations (MACs) and power consumption of hybrid ONNs using our polychromatic ONN as an example (details in the Supplementary). The total number of MAC operations in the simulated digital network is approximately 239 MMACs (while the full U-Net requires 65.9 GMACs and Efficient U-Net reaches 1.37 GMACs), which is reduced to 65 MMACs after incorporating optical frontends~\cite{Tan2019EfficientNetRM}. The total energy consumption includes both image capture and digital computation. The full U-Net (pre-trained teacher) consumes 2.03\,J for computation and 2.36\,mJ for image capture, totaling \textbf{2.04\,J} per image. The compact digital network requires 7.37\,mJ for computation and 2.36\,mJ for image capture, totaling \textbf{9.73\,mJ} per image. In contrast, our hybrid ONN consumes 3.82\,mJ for image capture and 2.01\,mJ for backend processing, totaling \textbf{5.83\,mJ}—representing over a 40\% reduction in system-level energy consumption compared to the simulated digital network, and over 300$\times$ energy compression compared to the pre-trained teacher U-Net.

\section{Conclusion}

We propose a comprehensive NTKD pipeline that addresses multiple tasks and multiple optical systems in ONN design, training, and compensation. By incorporating knowledge transfer, particularly NTK-based knowledge distillation, our framework consistently improves the accuracy of different optical systems across both classification and segmentation tasks. Based on extensive experiments, we observe that the current ONN performance is primarily limited by the shallow and linear nature of existing optical architectures. Future advances in deeper, nonlinear ONNs may help narrow the performance gap between optical and electronic neural networks, improving scalability and accuracy.

\bibliographystyle{unsrt}
\bibliography{refs}

\begin{thebibliography}{10}

\bibitem{xu2021survey}
Runqin Xu, Pin Lv, Fanjiang Xu, and Yishi Shi.
\newblock A survey of approaches for implementing optical neural networks.
\newblock {\em Optics \& Laser Technology}, 136:106787, 2021.

\bibitem{zhao2024towards}
Guibin Zhao, Pengfei Li, Zhibo Zhang, Fusen Guo, Xueting Huang, Wei Xu, Jinyin Wang, and Jianlong Chen.
\newblock Towards sar automatic target recognition: Multi-category sar image classification based on light weight vision transformer.
\newblock In {\em 2024 21st Annual International Conference on Privacy, Security and Trust (PST)}, pages 1--6. IEEE, 2024.

\bibitem{fu2024optical}
Tingzhao Fu, Jianfa Zhang, Run Sun, Yuyao Huang, Wei Xu, Sigang Yang, Zhihong Zhu, and Hongwei Chen.
\newblock Optical neural networks: progress and challenges.
\newblock {\em Light: Science \& Applications}, 13(1):263, 2024.

\bibitem{liu2024towards}
Yanbing Liu, Shaochong Liu, Tao Li, Tianyu Li, Wei Li, Guoqing Wang, Xun Liu, Wei Yang, and Yuan’an Liu.
\newblock Towards constructing a doe-based practical optical neural system for ship recognition in remote sensing images.
\newblock {\em Signal Processing}, 221:109488, 2024.

\bibitem{boloor2025privateeye}
Adith Boloor, Weikai Lin, Tianrui Ma, Yu~Feng, Yuhao Zhu, and Xuan Zhang.
\newblock Privateeye: In-sensor privacy preservation through optical feature separation.
\newblock In {\em 2025 IEEE/CVF Winter Conference on Applications of Computer Vision (WACV)}, pages 2357--2367. IEEE, 2025.

\bibitem{liu2025extrememeta}
Quan Liu, Brandon~T Swartz, Ivan Kravchenko, Jason~G Valentine, and Yuankai Huo.
\newblock Extrememeta: High-speed lightweight image segmentation model by remodeling multi-channel metamaterial imagers.
\newblock {\em Journal of Imaging Science and Technology}, pages 1--10, 2025.

\bibitem{zhao2023integrated}
Baiheng Zhao, Junwei Cheng, Bo~Wu, Dingshan Gao, Hailong Zhou, and Jianji Dong.
\newblock Integrated photonic convolution acceleration core for wearable devices.
\newblock {\em Opto-Electronic Science}, 2(12):230017--1, 2023.

\bibitem{colburn2019optical}
Shane Colburn, Yi~Chu, Eli Shilzerman, and Arka Majumdar.
\newblock Optical frontend for a convolutional neural network.
\newblock {\em Applied optics}, 58(12):3179--3186, 2019.

\bibitem{lu2017expressive}
Zhiyuan Lu, Huan Pu, Feicheng Wang, Zhiqiang Hu, and Liwei Wang.
\newblock The expressive power of neural networks: A view from the width.
\newblock In {\em NeurIPS}, 2017.

\bibitem{zhou2020universality}
Ding-Xuan Zhou.
\newblock Universality of deep convolutional neural networks.
\newblock {\em Applied and Computational Harmonic Analysis}, 48(2):787--794, 2020.

\bibitem{bao2021approximation}
Changshuo Bao, Qianxiao Li, and Haizhao Yang.
\newblock Approximation analysis of convolutional neural networks from a feature extraction view.
\newblock {\em Applied and Computational Harmonic Analysis}, 54:47--84, 2021.

\bibitem{han2015learning}
Song Han, Jeff Pool, John Tran, and William Dally.
\newblock Learning both weights and connections for efficient neural network.
\newblock {\em Advances in neural information processing systems}, 28, 2015.

\bibitem{molchanov2016pruning}
Pavlo Molchanov, Stephen Tyree, Tero Karras, Timo Aila, and Jan Kautz.
\newblock Pruning convolutional neural networks for resource efficient inference.
\newblock {\em arXiv preprint arXiv:1611.06440}, 2016.

\bibitem{seleznova2023neural}
Mariia Seleznova, Dana Weitzner, Raja Giryes, Gitta Kutyniok, and Hung-Hsu Chou.
\newblock Neural (tangent kernel) collapse.
\newblock {\em Advances in Neural Information Processing Systems}, 36:16240--16270, 2023.

\bibitem{liu2024physics}
Yanbing Liu, Jianwei Qin, Yan Liu, Xi~Yue, Xun Liu, Guoqing Wang, Tianyu Li, Ye~Ye, and Wei Li.
\newblock Physics-constrained comprehensive optical neural networks.
\newblock {\em Advances in Neural Information Processing Systems}, 37:92036--92054, 2024.

\bibitem{hu2021distilling}
Xinting Hu, Kaihua Tang, Chunyan Miao, Xian-Sheng Hua, and Hanwang Zhang.
\newblock Distilling causal effect of data in class-incremental learning.
\newblock In {\em Proceedings of the IEEE/CVF Conference on Computer Vision and Pattern Recognition}, pages 3957--3966, 2021.

\bibitem{xiang2022knowledge}
Jinlin Xiang, Shane Colburn, Arka Majumdar, and Eli Shlizerman.
\newblock Knowledge distillation circumvents nonlinearity for optical convolutional neural networks.
\newblock {\em Applied Optics}, 61(9):2173--2183, 2022.

\bibitem{harutyunyan2023supervision}
Hrayr Harutyunyan, Ankit~Singh Rawat, Aditya~Krishna Menon, Seungyeon Kim, and Sanjiv Kumar.
\newblock Supervision complexity and its role in knowledge distillation.
\newblock {\em arXiv preprint arXiv:2301.12245}, 2023.

\bibitem{lin2018diffractive}
Xing Lin, Yair Rivenson, Nezih~T. Yardimci, Muhammed Veli, Yi~Luo, Mona Jarrahi, and Aydogan Ozcan.
\newblock All-optical machine learning using diffractive deep neural networks.
\newblock {\em Science}, 361(6406):1004--1008, 2018.

\bibitem{mengu20diffractive}
Deniz Mengu, Yi~Luo, Yair Rivenson, and Aydogan Ozcan.
\newblock Analysis of diffractive optical neural networks and their integration with electronic neural networks.
\newblock {\em IEEE Journal of Selected Topics in Quantum Electronics}, 26(1), 2020.

\bibitem{yang2023optical}
Mingwei Yang, Elizabeth Robertson, Luisa Esguerra, Kurt Busch, and Janik Wolters.
\newblock Optical convolutional neural network with atomic nonlinearity.
\newblock {\em Optics Express}, 31(10):16451--16459, 2023.

\bibitem{ryou2021free}
Albert Ryou, James Whitehead, Maksym Zhelyeznyakov, Paul Anderson, Cem Keskin, Michal Bajcsy, and Arka Majumdar.
\newblock Free-space optical neural network based on thermal atomic nonlinearity.
\newblock {\em Photonics Research}, 9(4):B128--B134, 2021.

\bibitem{wang2023image}
Tianyu Wang, Mandar~M Sohoni, Logan~G Wright, Martin~M Stein, Shi-Yuan Ma, Tatsuhiro Onodera, Maxwell~G Anderson, and Peter~L McMahon.
\newblock Image sensing with multilayer nonlinear optical neural networks.
\newblock {\em Nature Photonics}, 17(5):408--415, 2023.

\bibitem{zheng2022meta}
Hanyu Zheng, Quan Liu, You Zhou, Ivan~I Kravchenko, Yuankai Huo, and Jason Valentine.
\newblock Meta-optic accelerators for object classifiers.
\newblock {\em Science Advances}, 8(30):eabo6410, 2022.

\bibitem{zhou2021large}
Tiankuang Zhou, Xing Lin, Jiamin Wu, Yitong Chen, Hao Xie, Yipeng Li, Jingtao Fan, Huaqiang Wu, Lu~Fang, and Qionghai Dai.
\newblock Large-scale neuromorphic optoelectronic computing with a reconfigurable diffractive processing unit.
\newblock {\em Nature Photonics}, 15(5):367--373, 2021.

\bibitem{huang2024photonic}
Luocheng Huang, Quentin~AA Tanguy, Johannes~E Fr{\"o}ch, Saswata Mukherjee, Karl~F B{\"o}hringer, and Arka Majumdar.
\newblock Photonic advantage of optical encoders.
\newblock {\em Nanophotonics}, 13(7):1191--1196, 2024.

\bibitem{xue2024fully}
Zhiwei Xue, Tiankuang Zhou, Zhihao Xu, Shaoliang Yu, Qionghai Dai, and Lu~Fang.
\newblock Fully forward mode training for optical neural networks.
\newblock {\em Nature}, 632(8024):280--286, 2024.

\bibitem{choi2024transferable}
Minho Choi, Jinlin Xiang, Anna Wirth-Singh, Seung-Hwan Baek, Eli Shlizerman, and Arka Majumdar.
\newblock Transferable polychromatic optical encoder for neural networks.
\newblock {\em arXiv preprint arXiv:2411.02697}, 2024.

\bibitem{chang2018hybrid}
Julie Chang, Vincent Sitzmann, Xiong Dun, Wolfgang Heidrich, and Gordon Wetzstein.
\newblock Hybrid optical-electronic convolutional neural networks with optimized diffractive optics for image classification.
\newblock {\em Scientific reports}, 8(1):1--10, 2018.

\bibitem{burgos2021design}
Carlos Mauricio~Villegas Burgos, Tianqi Yang, Yuhao Zhu, and A~Nickolas Vamivakas.
\newblock Design framework for metasurface optics-based convolutional neural networks.
\newblock {\em Applied Optics}, 60(15):4356--4365, 2021.

\bibitem{hu2022high}
Zibo Hu, Shurui Li, Russell~LT Schwartz, Maria Solyanik-Gorgone, Mario Miscuglio, Puneet Gupta, and Volker~J Sorger.
\newblock High-throughput multichannel parallelized diffraction convolutional neural network accelerator.
\newblock {\em Laser \& Photonics Reviews}, 16(12):2200213, 2022.

\bibitem{He24pluggable}
Cong He, Dan Zhao, Fei Fan, Hongqiang Zhou, Xin~Li adn Yao~Li, Junjie Li, Fei Dong, Yin-Xiao Miao, Yongtian Wang, and Lingling Huang.
\newblock Pluggable multitask diffractive neural networks based on cascaded metasurfaces.
\newblock {\em Opto-Electronic Advances}, 7(23005), 2024.

\bibitem{wei2024spatially}
Kaixuan Wei, Xiao Li, Johannes Froech, Praneeth Chakravarthula, James Whitehead, Ethan Tseng, Arka Majumdar, and Felix Heide.
\newblock Spatially varying nanophotonic neural networks.
\newblock {\em Science Advances}, 10(45):eadp0391, 2024.

\bibitem{zheng2024multichannel}
Hanyu Zheng, Quan Liu, Ivan~I Kravchenko, Xiaomeng Zhang, Yuankai Huo, and Jason~G Valentine.
\newblock Multichannel meta-imagers for accelerating machine vision.
\newblock {\em Nature nanotechnology}, 19(4):471--478, 2024.

\bibitem{he2024pluggable}
Cong He, Dan Zhao, Fei Fan, Hongqiang Zhou, Xin Li, Yao Li, Junjie Li, Fei Dong, Yin-Xiao Miao, Yongtian Wang, et~al.
\newblock Pluggable multitask diffractive neural networks based on cascaded metasurfaces.
\newblock {\em Opto-Electron. Adv}, 7(2):230005, 2024.

\bibitem{wirth2025compressed}
Anna Wirth-Singh, Jinlin Xiang, Minho Choi, Johannes~E Fr{\"o}ch, Luocheng Huang, Shane Colburn, Eli Shlizerman, and Arka Majumdar.
\newblock Compressed meta-optical encoder for image classification.
\newblock {\em Advanced Photonics Nexus}, 4(2):026009--026009, 2025.

\bibitem{wirth2024compressed}
Anna Wirth-Singh, Jinlin Xiang, Minho Choi, Johannes Fr{\"o}ch, Luocheng Huang, Eli Shlizerman, and Arka Majumdar.
\newblock Compressed meta-optical encoder for image classification.
\newblock In {\em CLEO: Fundamental Science}, pages FF1J--1. Optica Publishing Group, 2024.

\bibitem{huo2023optical}
Yuchi Huo, Hujun Bao, Yifan Peng, Chen Gao, Wei Hua, Qing Yang, Haifeng Li, Rui Wang, and Sung-Eui Yoon.
\newblock Optical neural network via loose neuron array and functional learning.
\newblock {\em Nature Communications}, 14(1):2535, 2023.

\bibitem{rahman2023time}
Md~Sadman~Sakib Rahman and Aydogan Ozcan.
\newblock Time-lapse image classification using a diffractive neural network.
\newblock {\em Advanced Intelligent Systems}, 5(5):2200387, 2023.

\bibitem{lee2020finite}
Jaehoon Lee, Lechao Xiao, Samuel~S Schoenholz, Yasaman Bahri, Roman Novak, Jascha Sohl-Dickstein, and Jeffrey Pennington.
\newblock Finite versus infinite neural networks: An empirical study.
\newblock In {\em Advances in Neural Information Processing Systems}, volume~33, pages 15156--15172, 2020.

\bibitem{ji2020knowledge}
Guangda Ji and Zhanxing Zhu.
\newblock Knowledge distillation in wide neural networks: Risk bound, data efficiency and imperfect teacher.
\newblock {\em Advances in Neural Information Processing Systems}, 33:20823--20833, 2020.

\bibitem{morello2024exploring}
Yoann Morello, Emilie Gr{\'e}goire, and Sam Verboven.
\newblock Exploring task affinities through ntk alignment and early training dynamics in multi-task learning.
\newblock In {\em NeurIPS 2024 Workshop on Mathematics of Modern Machine Learning}, 2024.

\bibitem{zheng2023dual}
Ziyang Zheng, Zhengyang Duan, Hang Chen, Rui Yang, Sheng Gao, Haiou Zhang, Hongkai Xiong, and Xing Lin.
\newblock Dual adaptive training of photonic neural networks.
\newblock {\em Nature Machine Intelligence}, 5(10):1119--1129, 2023.

\bibitem{spall2022hybrid}
James Spall, Xianxin Guo, and Alexander~I Lvovsky.
\newblock Hybrid training of optical neural networks.
\newblock {\em Optica}, 9(7):803--811, 2022.

\bibitem{ruiz2020deep}
Hans-Christian Ruiz~Euler, Marcus~N Boon, Jochem~T Wildeboer, Bram van~de Ven, Tao Chen, Hajo Broersma, Peter~A Bobbert, and Wilfred~G van~der Wiel.
\newblock A deep-learning approach to realizing functionality in nanoelectronic devices.
\newblock {\em Nature nanotechnology}, 15(12):992--998, 2020.

\bibitem{zhang2024general}
Yubo Zhang, Rui Chen, Minho Choi, Johannes~E Fr{\"o}ch, and Arka Majumdar.
\newblock General image compression using random-psf metasurfaces and computational back-end.
\newblock In {\em 2024 Conference on Lasers and Electro-Optics (CLEO)}, pages 1--3. IEEE, 2024.

\bibitem{zhao2024frequency}
Mengran Zhao, Shitao Zhu, Die Li, Thomas Fromenteze, Mohsen Khalily, Xiaoming Chen, Vincent Fusco, and Okan Yurduseven.
\newblock Frequency-diverse bunching metasurface antenna for microwave computational imaging.
\newblock {\em IEEE Transactions on Antennas and Propagation}, 2024.

\bibitem{pors2016random}
Anders Pors, Fei Ding, Yiting Chen, Ilya~P Radko, and Sergey~I Bozhevolnyi.
\newblock Random-phase metasurfaces at optical wavelengths.
\newblock {\em Scientific reports}, 6(1):28448, 2016.

\bibitem{dupre2018design}
Matthieu Dupr{\'e}, Liyi Hsu, and Boubacar Kant{\'e}.
\newblock On the design of random metasurface based devices.
\newblock {\em Scientific reports}, 8(1):7162, 2018.

\bibitem{wray2020light}
Parker~R Wray and Harry~A Atwater.
\newblock Light--matter interactions in films of randomly distributed unidirectionally scattering dielectric nanoparticles.
\newblock {\em ACS Photonics}, 7(8):2105--2114, 2020.

\bibitem{jacot2018neural}
Arthur Jacot, Franck Gabriel, and Cl{\'e}ment Hongler.
\newblock Neural tangent kernel: Convergence and generalization in neural networks.
\newblock {\em Advances in neural information processing systems}, 31, 2018.

\bibitem{NEURIPS2019_dbc4d84b}
Sanjeev Arora, Simon~S Du, Wei Hu, Zhiyuan Li, Russ~R Salakhutdinov, and Ruosong Wang.
\newblock On exact computation with an infinitely wide neural net.
\newblock In H.~Wallach, H.~Larochelle, A.~Beygelzimer, F.~d\textquotesingle Alch\'{e}-Buc, E.~Fox, and R.~Garnett, editors, {\em Advances in Neural Information Processing Systems}, volume~32. Curran Associates, Inc., 2019.

\bibitem{lecun1998gradient}
Yann LeCun, L{\'e}on Bottou, Yoshua Bengio, and Patrick Haffner.
\newblock Gradient-based learning applied to document recognition.
\newblock {\em Proceedings of the IEEE}, 86(11):2278--2324, 1998.

\bibitem{krizhevsky2009learning}
Alex Krizhevsky.
\newblock Learning multiple layers of features from tiny images.
\newblock Technical report, University of Toronto, 2009.
\newblock Technical report.

\bibitem{Tan2019EfficientNetRM}
Mingxing Tan and Quoc~V. Le.
\newblock Efficientnet: Rethinking model scaling for convolutional neural networks.
\newblock {\em ArXiv}, abs/1905.11946, 2019.

\end{thebibliography}

\end{document}